\documentclass[letterpaper, 10 pt, conference]{ieeeconf}

\IEEEoverridecommandlockouts
\title{\LARGE \bf
    VAMPS: Visual and Motor Policies from Sampling-Based Planning
}

\author{\vspace{-0.3em}%
Mohamed Yassine Kabouri$^{1,2}$, Pietro Noah Crestaz$^{1,3}$,
Quang-Nam Nguyen$^{2}$,
Qilong Cheng$^{2}$,\\[0.35em] Ludovic Righetti$^{2,4}$, and Nicolas Mansard$^{1,4}$%
\thanks{$^{1}$LAAS-CNRS, Universit\'e de Toulouse, CNRS, Toulouse, France.}%
\thanks{$^{2}$Machines in Motion Laboratory, New York University, New York, USA.}%
\thanks{$^{3}$Industrial Engineering Department, University of Trento, Trento, Italy.}%
\thanks{$^{4}$Artificial and Natural Intelligence Toulouse Institute (ANITI), Toulouse.}}

\usepackage{amsmath,amsthm,amsfonts,amssymb,amscd, fancyhdr, color, comment, graphicx, environ}
\usepackage{hyperref}
\usepackage{array}
\usepackage{placeins}
\usepackage{tcolorbox}
\usepackage[noend]{algpseudocode}
\usepackage{algorithm}
\usepackage{graphicx}
\usepackage{subcaption}
\usepackage{xcolor}
\usepackage{wrapfig}
\usepackage[font=footnotesize,labelfont=bf,skip=6pt]{caption}
\newcommand{\E}{\mathbb{E}}
\makeatletter
\floatevery{algorithm}{%
  \setlength{\textfloatsep}{8pt plus 2pt minus 2pt}%
  \def\@fs@pre{\hrule width\dimexpr\columnwidth-1pt\relax
    height.8pt depth0pt\kern2pt}%
  \def\@fs@mid{\kern2pt\hrule width\dimexpr\columnwidth-1pt\relax
    \kern2pt}%
  \def\@fs@post{\kern2pt\hrule width\dimexpr\columnwidth-1pt\relax
    \vspace{-0.6em}}%
}
\makeatother

\newcommand{\figref}[1]{Fig.~\ref{#1}}

\newcommand{\secref}[1]{Sec.~\ref{#1}}

\theoremstyle{remark}

\IEEEaftertitletext{\vspace{-1.5\baselineskip}}

\usepackage[T1]{fontenc}
\usepackage{booktabs}

\begin{document}

\maketitle
\thispagestyle{empty}
\pagestyle{empty}


\begin{abstract}
Learning robot policies directly on physical systems remains difficult because data collection is costly and policy exploration can be unsafe. We introduce Visual and Motor Policies from Sampling-Based Planning (VAMPS), a framework that uses Model Predictive Path Integral (MPPI) control to train reusable policies without human demonstrations. VAMPS supports two training modes. For one-step proprioceptive policies, it operates iteratively in simulation: the policy warm-starts MPPI, and the refined trajectories provide new supervision as the policy changes. A learned terminal value improves short-horizon planning, while an Implicit Q-Learning (IQL) critic guides the policy update. Iterative refinement outperforms training once on frozen MPPI data, and we transfer the learned locomotion policy to a Unitree Go2. For visuomotor policies, VAMPS operates directly from real-robot data. MPPI uses task-specific state estimates to plan and execute trajectories while recording RGB and sensor observations on a Flexiv Rizon 10S. Action Chunking with Transformers predicts action chunks, reducing the effective prediction horizon, and is trained offline on this fixed dataset. We demonstrate visuomotor pick-and-place and force-aware whiteboard erasing. In the latter task, the policy additionally observes the measured $6$-D wrench and desired normal force. These results show that VAMPS can learn policies either in simulation followed by hardware transfer or directly from autonomously collected real-robot data.
\end{abstract}
%

\section{Introduction}

Learning directly on physical robots, through interaction and sensor feedback, remains a major challenge, despite being the most direct route to acquiring effective sensorimotor behaviors. This is especially compelling for contact-rich robotics, including pick-and-place manipulation, force-regulated interaction, and quadruped locomotion. Current learning methods do not yet solve this problem at scale: they typically rely on human demonstrations or move exploration to simulation before transferring the resulting policy to hardware. This work takes a step toward methods that can autonomously and efficiently collect sensorimotor experience on robots and turn it into better policies.

This objective has been pursued through three main routes. \textbf{Imitation learning} methods such as Action Chunking with Transformers (ACT)~\cite{Zhao-RSS-23} and Diffusion Policy~\cite{Chi2023DiffusionPV} clone human teleoperation and learn rich behaviors. Their bottleneck is data: demonstrations are expensive, performance is bounded by the operator, and cloning a fixed dataset leads to compounding errors~\cite{dagger_2011}. DAgger~\cite{dagger_2011} corrects those states by querying an expert policy on states visited by the learned policy. \textbf{Reinforcement learning}~\cite{Hwangbo2019LearningAA} needs no demonstration, but it has to explore, and exploration is a bottleneck on contact-rich tasks with short episodes~\cite{mao2025learning}. 
\textbf{Trajectory Optimization} (TO)~\cite{Zhao_MOTO,Jordana_mim_solver}, in contrast, optimizes individual rollouts for a given cost. While gradient-based TO relies on derivatives that are challenging to compute in contact-rich tasks, sampling-based planners~\cite{williams2017informationtheoreticmodelpredictive} handle non-smooth contact dynamics and run in parallel on GPUs~\cite{Xue2024FullOrderSM}. Yet their search space scales exponentially with the planning horizon~\cite{pnc_td_cd_mppi}.
They are commonly used in receding horizon, which requires online replanning and does not directly produce a reusable policy.

\begin{figure*}[!t]
  \centering
  \includegraphics[width=\textwidth,trim=3 4 3 6,clip]{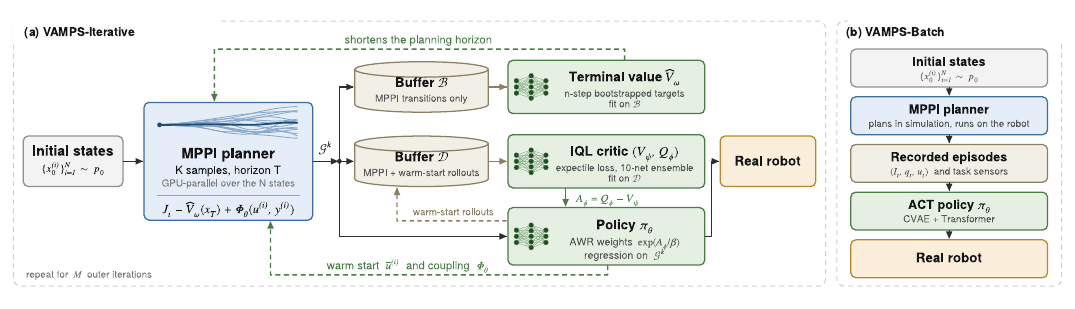}
  \vspace{-1.6em}
    \caption{Overview of the two VAMPS modes. \textbf{(a)} \textit{VAMPS-Iterative:} a one-step proprioceptive policy warm-starts MPPI, which refines its behavior and provides new supervision. A terminal value improves short-horizon planning, and an IQL critic weights the policy updates. \textbf{(b)} \textit{VAMPS-Batch:} MPPI uses task-specific estimates to autonomously collect RGB, sensor and action episodes on the real robot, and ACT trains offline on the resulting fixed dataset.}
  \vspace{-0.7em}
  \label{fig:teaser}
\end{figure*}

The complementarity between trajectory optimization and policy learning has long been recognized, notably by Guided Policy Search (GPS)~\cite{gpslevine} and subsequent optimizer-policy coupling methods~\cite{mordatch_todo, lidec2023enforcingconsensustrajectoryoptimization}. This perspective combines the precision of model-based optimization with the fast execution of a learned controller, while avoiding direct exploration by the learned policy. We build on this principle and use MPPI as a behavior generator from a model and a task cost.

In this work, we introduce Visual and Motor Policies from Sampling-Based Planning (VAMPS), a framework that distills trajectories generated by MPPI into reusable robot policies without human demonstrations or simulator gradients. MPPI plans with a model, a task cost and task-specific state estimates, while the learned policy maps the observations available at deployment directly to actions. VAMPS supports two training modes that share this planner-to-policy principle.

\textbf{VAMPS-Iterative} trains a one-step proprioceptive policy in simulation. Small prediction errors can move such a policy away from the states in a fixed training dataset and compound over time~\cite{dagger_2011}. The policy therefore repeatedly warm-starts MPPI, which refines its behavior and provides new supervision as the policy evolves. This loop does not require simulator gradients. A learned terminal value improves short-horizon planning, while an IQL~\cite{kostrikov2022offline} critic guides the policy update through Advantage-Weighted Regression (AWR)~\cite{peng2021advantageweighted}. We train the policy in simulation and transfer it to a real robot.

\textbf{VAMPS-Batch} uses MPPI to autonomously collect a real-robot dataset and train a visuomotor policy from it. MPPI uses task-specific state estimates to plan and execute trajectories on the robot while RGB images, robot sensors and actions are recorded. ACT~\cite{Zhao-RSS-23} predicts action chunks, reducing the effective prediction horizon. For the tasks considered here, we collect the MPPI episodes once, freeze the dataset and train ACT offline. At deployment, ACT maps images and robot sensors directly to action chunks and does not use the task-specific state estimates.

Our contributions are threefold. (1) We introduce \textbf{V}isual \textbf{a}nd \textbf{M}otor \textbf{P}olicies from \textbf{S}ampling-Based Planning (\textbf{VAMPS}), a framework that uses MPPI to train reusable policies either in simulation or directly from real-robot data. (2) VAMPS-Iterative trains one-step proprioceptive policies. We show that iterative refinement outperforms training once on frozen MPPI data, a simple behavioral-cloning coupling between MPPI and the policy is sufficient, and a learned terminal value improves short-horizon planning. We transfer the resulting locomotion policy to a Unitree Go2. (3) VAMPS-Batch uses MPPI to autonomously collect a fixed real-robot dataset without human demonstrations and trains chunked ACT policies on it. We demonstrate this mode on pick-and-place and force-aware whiteboard erasing with a Flexiv Rizon 10S.

\section{Related Work}\label{sec:related_work}

GPS~\cite{gpslevine} and its visuomotor extension~\cite{levine2016endtoendtrainingdeepvisuomotor} pioneered the idea of using local TO controllers to supervise policy learning. In the same cycle of ideas, ~\cite{mordatch_todo} formalized the TO-policy coupling via ADMM to enforce agreement between them. ~\cite{lidec2023enforcingconsensustrajectoryoptimization} later used augmented-Lagrangian updates and sensitivity information to improve this coupling. CACTO-SL instead combines TO with actor-critic learning and uses the policy to warm-start the optimizer~\cite{alboni2023cactoslusingsobolevlearning}. More recently, ~\cite{lehellard2026accelerating} trained diffusion policies to warm-start gradient-based TO. Their method alternates policy learning and trajectory refinement, and uses the feedback gains returned by the optimizer in the training loss. These methods rely mainly on gradient-based TO and differentiable dynamics.

PI$^2$GPS~\cite{Chebotar2016PathIG} replaces the gradient-based local optimizer with the derivative-free PI$^2$ algorithm. However, it initializes the local policies from kinesthetic demonstrations and pretrains its visual features on a pose-prediction task. This work studies how to scale the GPS idea using modern tools from derivative-free optimization~\cite{jordana2025introductionzeroorderoptimizationtechniques}, offline RL~\cite{Levine2020OfflineRL} and imitation learning~\cite{Zhao-RSS-23}. VAMPS-Iterative uses GPU-parallel MPPI in receding horizon. It requires a model and a task cost but does not require simulator derivatives, feedback gains or a human warm start. The policy repeatedly warm-starts MPPI, which refines its behavior and provides new supervision. To improve short-horizon planning, we introduce a learned terminal value trained on the planner transitions. To filter the planner actions, we train an IQL critic~\cite{kostrikov2022offline} on MPPI trajectories and policy rollouts, and use its advantage in AWR~\cite{peng2021advantageweighted} to give more weight to high-advantage actions.

Visuomotor imitation methods such as ACT~\cite{Zhao-RSS-23} and Diffusion Policy~\cite{Chi2023DiffusionPV} predict sequences of actions from visual observations. Action-sequence prediction reduces the effective prediction horizon and limits the accumulation of one-step errors. These methods are commonly trained from human demonstrations. VAMPS-Batch instead uses MPPI to collect real-robot RGB, sensor and action episodes, and trains ACT on the resulting fixed dataset. Task-specific state estimates guide MPPI during data collection but are not provided to the deployed policy. For force-aware erasing, MPPI records RGB images, the measured wrench and the desired normal force, and ACT learns a force-conditioned policy from these episodes. We deploy the proprioceptive policy learned by VAMPS-Iterative on a Unitree Go2 and the visuomotor policies learned by VAMPS-Batch on a Flexiv Rizon 10S.

\section{Preliminaries}\label{sec:preliminaries}

\begin{figure*}[!t]
  \vspace*{7pt}
  \centering
  \includegraphics[width=0.238\textwidth]{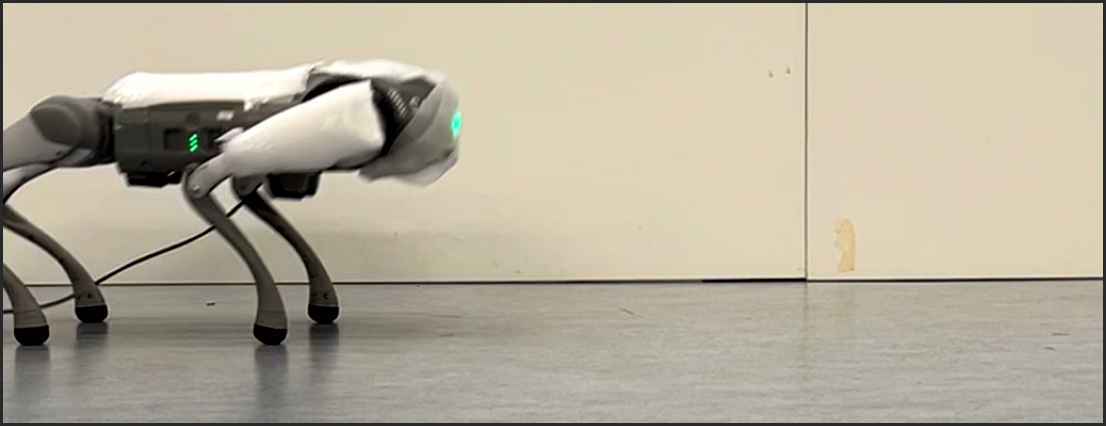}\hspace{1.8pt}%
  \includegraphics[width=0.238\textwidth]{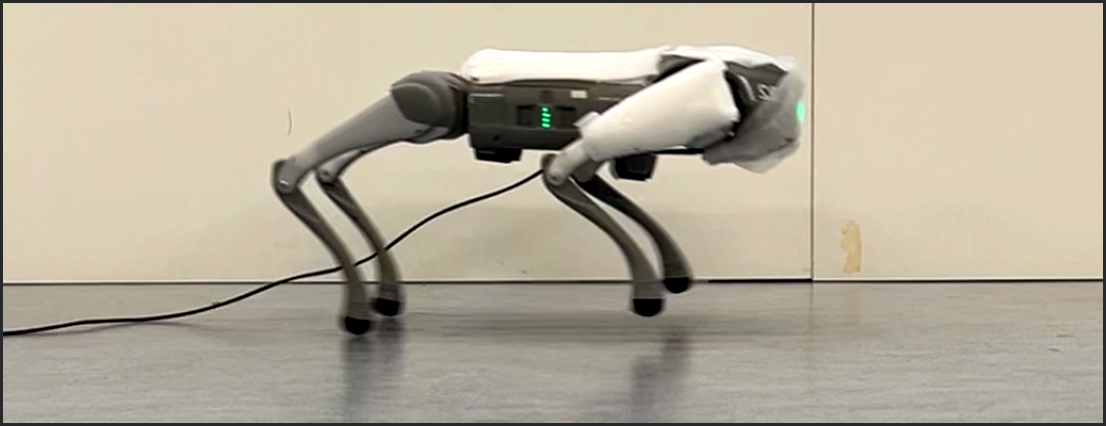}\hspace{1.8pt}%
  \includegraphics[width=0.238\textwidth]{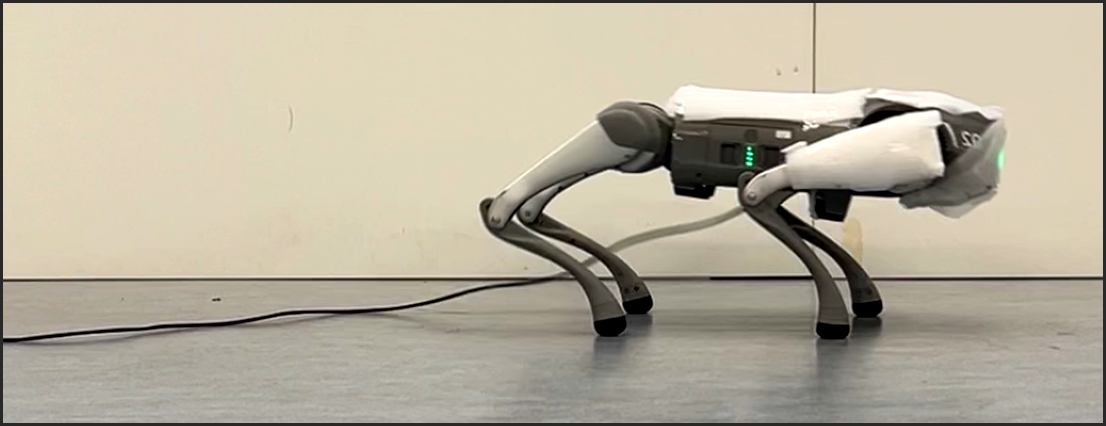}\hspace{1.8pt}%
  \includegraphics[width=0.238\textwidth]{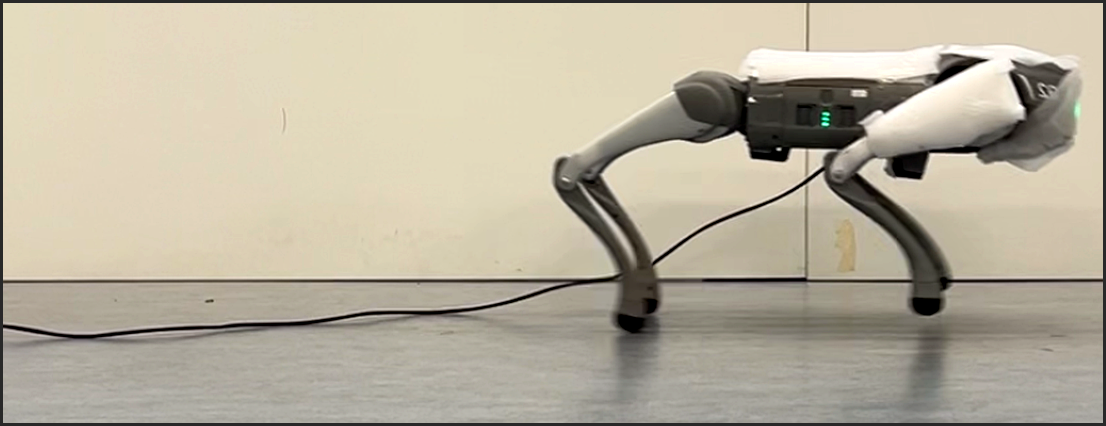}\\[1.6pt]
  \includegraphics[width=0.238\textwidth]{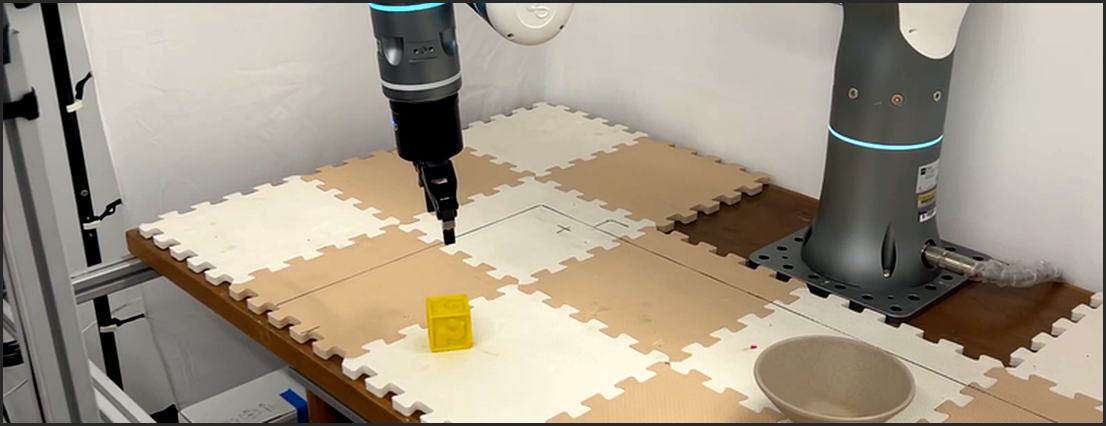}\hspace{1.8pt}%
  \includegraphics[width=0.238\textwidth]{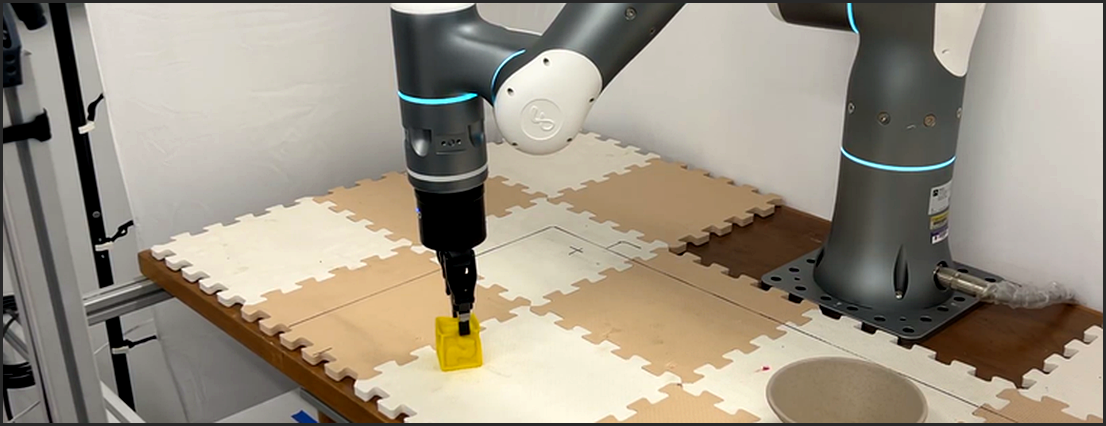}\hspace{1.8pt}%
  \includegraphics[width=0.238\textwidth]{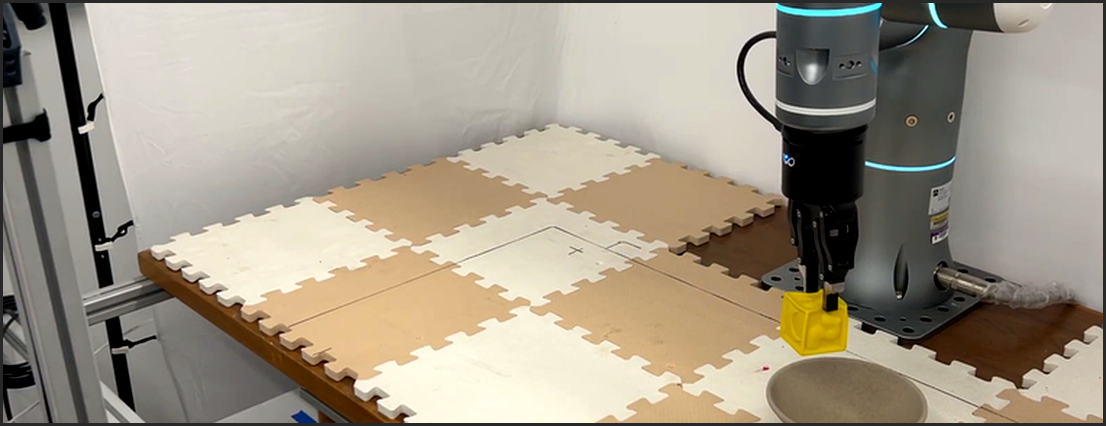}\hspace{1.8pt}%
  \includegraphics[width=0.238\textwidth]{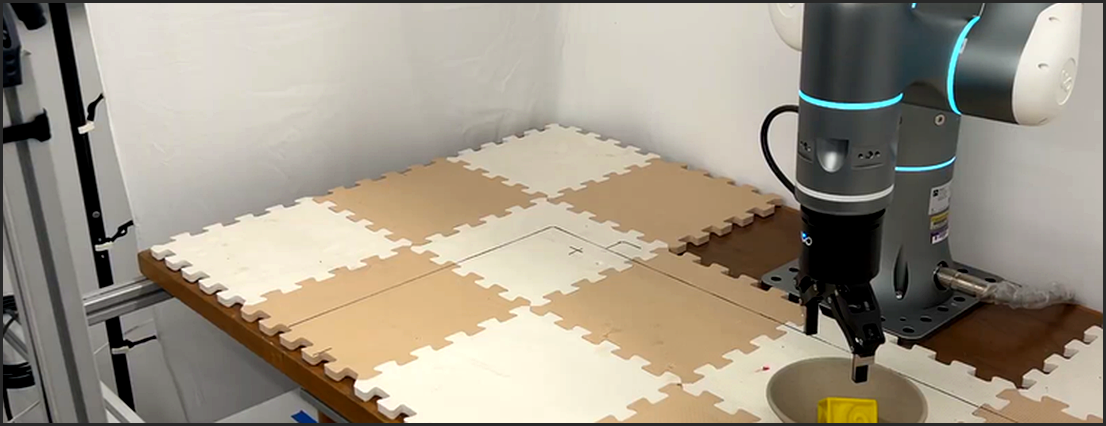}\\[1.6pt]
  \includegraphics[width=0.238\textwidth]{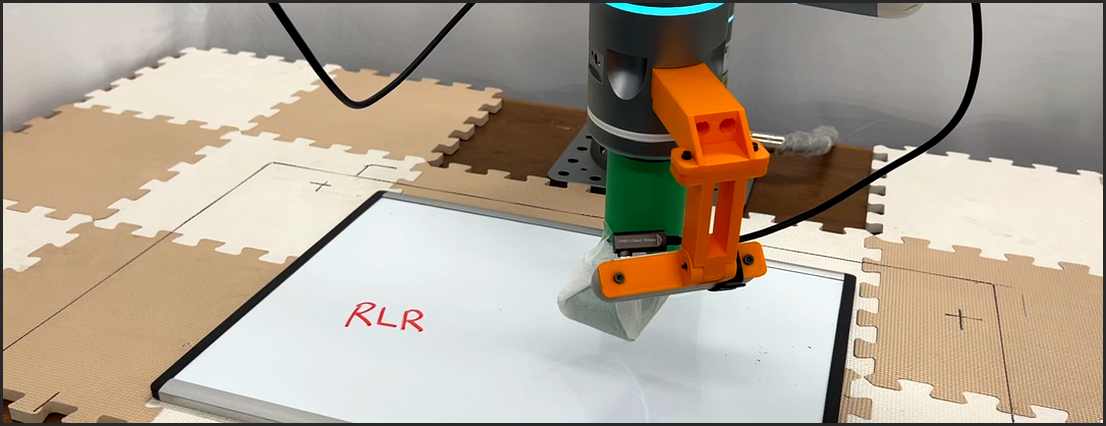}\hspace{1.8pt}%
  \includegraphics[width=0.238\textwidth]{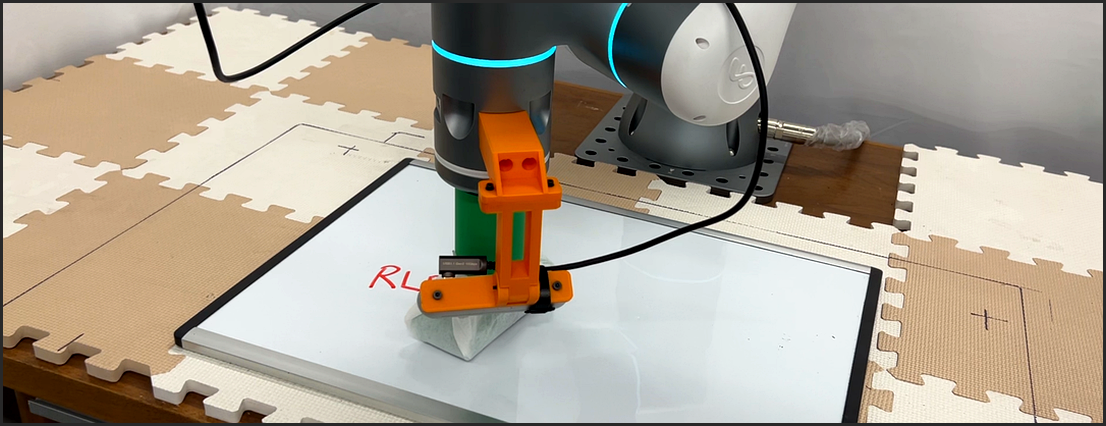}\hspace{1.8pt}%
  \includegraphics[width=0.238\textwidth]{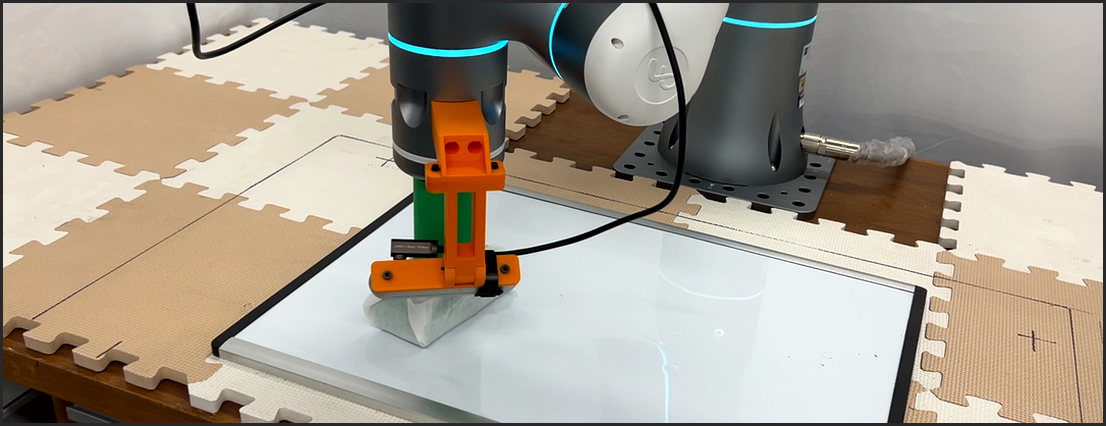}\hspace{1.8pt}%
  \includegraphics[width=0.238\textwidth]{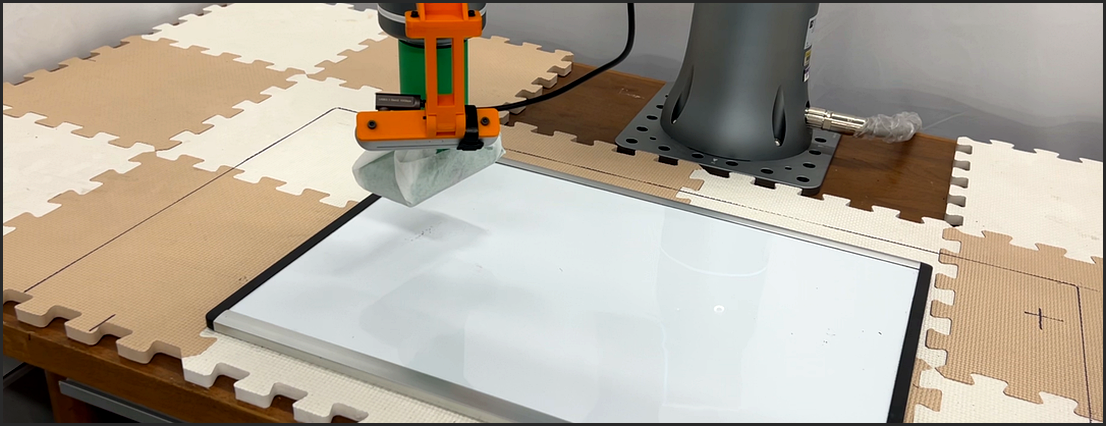}
  \caption{The hardware that we consider, four frames each in temporal order. Top: bounding policy of the Unitree Go2 under a forward velocity command, with a proprioceptive policy trained in simulation. Middle: pick-and-place on the Flexiv Rizon 10S. Bottom: whiteboard erasing on the same arm. The two arm tasks are visuomotor and are learned directly from hardware data. Videos are provided in the supplementary video.}
  \vspace{-0.7em}
  \label{fig:go2_deploy}
\end{figure*}

\subsection{Model Predictive Path Integral (MPPI)}\label{sec:mppi}
MPPI~\cite{williams2017informationtheoreticmodelpredictive} is a sampling-based trajectory optimization method. From the current state $\mathbf{x}_0$, it solves the following finite-horizon optimal control problem (OCP):
\begin{equation}\label{eq:J}
\begin{aligned}
    \min_{\mathbf{u}_{0:T-1}} \quad & J(\mathbf{x}_0,\,\mathbf{u}) = \sum_{t=0}^{T-1} \gamma^{t} c\bigl(\mathbf{x}_t,\,\mathbf{u}_t\bigr) + \gamma^{T} c_f(\mathbf{x}_T)\\
    \text{s.t.} \quad & \mathbf{x}_{t+1} = f(\mathbf{x}_t, \mathbf{u}_t), \quad \forall t = 0, \dots, T-1, \\
    & \mathbf{u}_t \in \mathcal{U}, \quad \mathbf{x}_t \in \mathcal{X},
\end{aligned}
\end{equation}
Here, $\mathbf{u}=(\mathbf{u}_0,\dots,\mathbf{u}_{T-1})$ and $\gamma$ is the discount factor. We use the cost convention $c=-r$ in the ocp, while the experiments report return, for which higher values are better.

MPPI samples $K$ perturbations $w^{(i)}\sim\mathcal{N}(0,\Sigma_{0:T-1})$ around the nominal control sequence $\mathbf{u}$. The cost of each perturbed trajectory is $J(\mathbf{x}_0,\mathbf{u}+w^{(i)})$, where the states are obtained by rolling out the dynamics from $\mathbf{x}_0$. This cost gives the standard MPPI update:
\begin{equation}
\begin{aligned}
\mathbf{u} \gets \mathbf{u} + \frac{\sum_{i=1}^K \exp\Big(-\tfrac{1}{\lambda} J(\mathbf{x}_0,\mathbf{u}+w^{(i)})\Big) \, w^{(i)}}
{\sum_{i=1}^K \exp\Big(-\tfrac{1}{\lambda} J(\mathbf{x}_0,\mathbf{u}+w^{(i)})\Big)}.
\end{aligned}
\end{equation}

We run MPPI in a receding-horizon loop. A short horizon of the optimized control sequence is applied to the system, after which the control sequence is shifted and MPPI solves a new OCP from the resulting state.

\subsection{Implicit Q-Learning}\label{sec:iql_prelim}
IQL~\cite{kostrikov2022offline} learns a state value $V_\psi$ and an action value $Q_\phi$ from a dataset of transitions $\mathcal{D}$. It avoids maximizing $Q_\phi$ over actions outside the dataset, which can cause value overestimation~\cite{Levine2020OfflineRL}. Instead, it approximates the in-sample maximum with the asymmetric squared expectile loss $L_2^\tau$:
\begin{align}
  \mathcal{L}_V(\psi) &=
    \E_{(x,u)\sim\mathcal{D}}\!\bigl[
      L_2^\tau\!\bigl(Q_\phi(x,u) - V_\psi(x)\bigr)
    \bigr], \label{eq:iql_v}\\
  \mathcal{L}_Q(\phi) &=
    \E_{\mathcal{D}}\!\bigl[
      \bigl(r + \gamma V_\psi(x') - Q_\phi(x,u)\bigr)^{\!2}
    \bigr], \label{eq:iql_q}
\end{align}
where $L_2^\tau(e)=|\tau-\mathbf{1}[e<0]|e^2$. The resulting advantage is $A_\phi(x,u)=Q_\phi(x,u)-V_\psi(x)$.

\subsection{Action Chunking with Transformers (ACT)}\label{sec:act}

ACT~\cite{Zhao-RSS-23} is a visuomotor policy that predicts a short action trajectory (called a chunk) of $T_u$ consecutive joint-position targets instead of one action. This reduces the accumulation of prediction errors.

Let $I_{t,c}$ denote the RGB image from camera $c$, $\mathbf{q}_t$ the robot sensor observation, and $F_{t,c}=\operatorname{ResNet}(I_{t,c})$ the spatial image features. ACT is trained as a conditional variational autoencoder (CVAE). Its policy uses a Transformer encoder to combine $(\mathbf{q}_t,\{F_{t,c}\},\mathbf{z})$, and a Transformer decoder predicts $\hat{\mathbf{u}}_{t:t+T_u-1}$. The latent variable $\mathbf{z}$ represents the variation between action trajectories. During training, a separate CVAE encoder infers $\mathbf{z}$ from $(\mathbf{q}_t,\mathbf{u}_{t:t+T_u-1})$. Training minimizes
\begin{equation}\label{eq:act_loss}
\begin{split}
    \mathcal{L} = {}& \mathbb{E}_{\mathbf{z}\sim q_\phi}\bigl[\|\mathbf{u}_{t:t+T_u-1} - \hat{\mathbf{u}}_{t:t+T_u-1}\|_1\bigr] \\
    &+ \beta_{\mathrm{KL}}\, D_{\mathrm{KL}}\!\left(q_\phi(\mathbf{z}\mid\mathbf{q}_t,\mathbf{u}_{t:t+T_u-1})\,\|\,\mathcal{N}(\mathbf{0},\mathbf{I})\right).
\end{split}
\end{equation}
At inference the CVAE encoder is discarded and $\mathbf{z}=\mathbf{0}$, the mean of the prior.

\section{VAMPS: Learning Policies from Sampling-Based Planning}\label{sec:method}

VAMPS uses MPPI in two complementary modes. VAMPS-Iterative alternates between MPPI trajectory refinement and policy fitting, whereas VAMPS-Batch first collects a fixed dataset with MPPI and then trains a policy offline.

\subsection{VAMPS-Iterative}\label{sec:vamps_iterative}
Our goal is to learn a policy $\pi_\theta$ over a distribution of initial states $p_0$. Rolling out $\pi_\theta$ from $\mathbf{x}_0$ produces a control sequence whose cost is $J$ in Equation ~\eqref{eq:J}; we write $J^{\pi_\theta}(\mathbf{x}_0)$ for that cost. The policy search problem is
\begin{equation}\label{eq:policy_obj}
  \min_{\theta}\;
  \mathbb{E}_{\mathbf{x}_0\sim p_0}\bigl[J^{\pi_\theta}(\mathbf{x}_0)\bigr].
\end{equation}
In practice, we sample a finite set of initial states $\{x_0^{(i)}\}_{i=1}^N \sim p_0$. Following GPS~\cite{NIPS2016_a00e5eb0} and~\cite{lidec2023enforcingconsensustrajectoryoptimization}, we introduce a control sequence $\mathbf{u}^{(i)}$ for each sampled initial state and require these controls to be reproducible by the policy:
\begin{equation}
\label{eq:gps_constrained}
\begin{aligned}
  \min_{\theta,\,\{\mathbf{u}^{(i)}\}_{i=1}^N} \quad
    & \frac{1}{N}\sum_{i=1}^{N} J\!\bigl(x_0^{(i)},\,\mathbf{u}^{(i)}\bigr) \\
  \text{s.t.} \quad
    & \mathbf{x}_{t+1}^{(i)} = f\bigl(x_t^{(i)},\,u_t^{(i)}\bigr), \\
    & u_t^{(i)} = \pi_\theta\!\bigl(x_t^{(i)}\bigr),
      \quad \forall\, i = 1,\dots,N, \;\forall\, t.
\end{aligned}
\end{equation}
We consider three strategies for handling this constraint in \secref{sec:coupling}. For a Gaussian policy, we express the constraint as a KL divergence between the MPPI trajectory distribution and the policy distribution.

\subsubsection{Policy-Guided MPPI with a Learned Terminal Value}\label{sec:mppi_step}
VAMPS-Iterative alternates between trajectory refinement with MPPI and policy fitting. At outer iteration $k$, we run MPPI in receding-horizon fashion from each sampled initial state $x_0^{(i)}$. At every replanning step, the current policy provides the nominal control sequence by rolling out the dynamics,
\begin{equation}\label{eq:warm_start}
  \bar{u}_t^{(i)}=\pi_\theta(x_t^{(i)}),\quad x_{t+1}^{(i)}=f(x_t^{(i)},\bar{u}_t^{(i)}),
\end{equation}
and we write $\bar{\mathbf{u}}^{(i)}=(\bar{u}_0^{(i)},\ldots,\bar{u}_{T-1}^{(i)})$. MPPI refines this sequence, and the refined controls $\mathbf{u}^{(i)}$ supervise the policy. We extend the effective planning horizon through bootstrapping~\cite{pnc_td_cd_mppi} by setting the terminal cost in~\eqref{eq:J} to $c_f(x_T)=-\hat{V}_\omega(x_T)$. The minus sign converts the predicted return into a cost. The value function is learned from MPPI transitions by regression onto $n$-step bootstrapped returns:
\begin{equation}\label{eq:terminal_v}
\begin{aligned}
  \mathcal{L}_{\hat{V}}(\omega) &=
    \E_{(x_t,\,G_t)\sim\mathcal{B}}\!\Bigl[
      \bigl(\hat{V}_\omega(x_t) - G_t\bigr)^{2}
    \Bigr], \\
  G_t &= \sum_{j=0}^{n-1}\gamma^{j} r_{t+j}
        + \gamma^{n}\hat{V}_{\omega^-}(x_{t+n}),
\end{aligned}
\end{equation}
with $\gamma$ the discount factor, $\omega^-$ the parameters of a target network updated by Polyak averaging, and $\mathcal{B}$ a dedicated buffer holding only $(x_t, G_t)$ pairs from MPPI trajectories.

\begin{algorithm}[t]
\caption{VAMPS-Iterative}
\label{alg:vamps_iterative}
\begin{algorithmic}[1]
\Require initial-state distribution $p_0$, number of states $N$, horizon $T$,
         outer iterations $M$, penalty $\rho$, coupling $\in\{\text{BC},\,\text{Penalty},\,\text{ADMM}\}$
\State Sample $\{x_0^{(i)}\}_{i=1}^N \sim p_0$
       \Comment{once, fixed for all iterations}
\State Initialize policy $\theta$, terminal value $\omega$, critic $(\psi,\phi)$
\State $\mathbf{y}^{(i)}\!\leftarrow\mathbf{0}$,\;
       $\mathcal{D}\leftarrow\emptyset$,\;
       $\mathcal{B}\leftarrow\emptyset$
\For{$k = 1$ \textbf{to} $M$}
  \For{$i = 1$ \textbf{to} $N$} \Comment{GPU-parallel}
    \State $\bigl(\bar{x}^{(i)}_t,\, \bar{u}^{(i)}_t\bigr)_{t=0}^{T-1}
           \leftarrow \text{rollout } \pi_\theta$
           \Comment{warm start}
    \State $\mathbf{u}^{(i)} \leftarrow
           \mathrm{MPPI}\bigl(J_i
           + \Phi_\theta(\mathbf{u}^{(i)},\mathbf{y}^{(i)})\bigr)$
           \label{line:mppi}
           \Comment{$c_f=-\hat{V}_\omega$}
  \EndFor
  \State $\mathcal{G}^k \leftarrow \{(x_t^{(i)}, u_t^{(i)})\}_{i,t}$
  \State $\mathcal{B} \leftarrow \mathcal{B} \cup \mathcal{G}^k$
  \State $\mathcal{D} \leftarrow \mathcal{D} \cup \mathcal{G}^k
         \cup \bigl\{(\bar{x}^{(i)}_t,\, \bar{u}^{(i)}_t)\bigr\}_{i,t}$
  \State Update $\hat{V}_\omega$ on $\mathcal{B}$
         \Comment{\eqref{eq:terminal_v}}
  \State Update $V_\psi,\, Q_\phi$ on $\mathcal{D}$ by IQL
         \Comment{\eqref{eq:iql_v}, \eqref{eq:iql_q}}
  \State Update $\theta$ on $\mathcal{G}^k$
         \Comment{\eqref{eq:vamps_policy_update}}
  \If{ADMM}
    \State $y_t^{(i)} \leftarrow
           y_t^{(i)} + u_t^{(i)} - \mu_\theta(x_t^{(i)}),\quad \forall i,t$
           \Comment{\eqref{eq:admm_dual}}
  \EndIf
\EndFor
\State \Return $\theta$
\end{algorithmic}
\end{algorithm}

\subsubsection{Planner-Policy Coupling}\label{sec:coupling}
VAMPS-Iterative supports three coupling strategies: BC, Penalty and ADMM. They differ in how they enforce the constraint in~\eqref{eq:gps_constrained}. We write $J_i(\mathbf{u})=J(x_0^{(i)},\mathbf{u})$ and denote the control sequence returned by MPPI as $\mathrm{MPPI}(J_i)$.

We measure the disagreement between a planner action and the policy with
\begin{equation}\label{eq:coupling_discrepancy}
d_\theta(x,u,y)=
\begin{cases}
\|u+y-\pi_\theta(x)\|^2, & \text{deterministic},\\
D_\text{KL}\!\left(q_{u+y}\,\|\,\pi_\theta(\cdot\mid x)\right), & \text{Gaussian},
\end{cases}
\end{equation}
where $q_v=\mathcal{N}(v,\sigma_\text{MPPI}^2\mathbf{I})$ is the MPPI trajectory distribution for a Gaussian policy~\cite{element_info_theory}. In ADMM, $y=\lambda/\rho\in\mathbb{R}^{n_u}$ is the scaled dual variable associated with the planner-policy consensus constraint, where $\lambda$ is the corresponding Lagrange multiplier and $\rho>0$ is the penalty parameter. We define the trajectory-level coupling penalty as
\begin{equation}\label{eq:coupling_penalty}
\Phi_\theta\!\left(\mathbf{u}^{(i)},\mathbf{y}^{(i)}\right)
=\frac{\rho}{2}\sum_t
d_\theta\!\left(x_t^{(i)},u_t^{(i)},y_t^{(i)}\right).
\end{equation}
Here, $\rho>0$ is the coupling coefficient.

\textbf{BC coupling.} The planner is uncoupled from the current policy and runs MPPI on the task cost alone:
\begin{equation}\label{eq:bc_traj}
  \mathbf{u}^{(i),k+1} \leftarrow \mathrm{MPPI}(J_i).
\end{equation}

\textbf{Penalty coupling.} The coupling $\Phi_\theta$ (with $\mathbf{y}^{(i)} = \mathbf{0}$) is added to the MPPI cost:
\begin{equation}\label{eq:penalty_traj}
  \mathbf{u}^{(i),k+1} \leftarrow
  \mathrm{MPPI}\!\left(J_i + \Phi_\theta\!\left(\mathbf{u}^{(i)}, \mathbf{0}\right)\right).
\end{equation}
The penalty biases MPPI toward trajectories the current policy can represent.

\textbf{ADMM coupling.} The constraint of~\eqref{eq:gps_constrained} is the consensus $\mathbf{u}^{(i)}=\bar{\mathbf{u}}^{(i)}$. Using the scaled dual variables $\mathbf{y}^{(i)}$, the augmented Lagrangian is:
\begin{equation}\label{eq:aug_lagrangian}
  \mathcal{L}_\rho =
    \sum_{i=1}^N J_i
    + \sum_{i=1}^N \Phi_\theta\!\left(\mathbf{u}^{(i)}, \mathbf{y}^{(i)}\right).
\end{equation}
Its trajectory update is
\begin{equation}\label{eq:admm_traj}
  \mathbf{u}^{(i),k+1} \leftarrow
  \mathrm{MPPI}\!\left(J_i + \Phi_\theta\!\left(\mathbf{u}^{(i)},\, \mathbf{y}^{(i),k}\right)\right).
\end{equation}

\begin{algorithm}[t]
\caption{VAMPS-Batch}
\label{alg:vamps_batch}
\begin{algorithmic}[1]
\Require task cost $J$, initial-state distribution $p_0$, episodes $E$
\State $\mathcal{D}_\text{batch}\leftarrow\emptyset$
\For{$i=1$ \textbf{to} $E$}
  \State Sample $x_0^{(i)}\sim p_0$
  \Statex \hspace{\algorithmicindent}\textit{\# Replan online and record data}
  \State $\tau_i\leftarrow\mathrm{ExecuteOnRobot}\bigl(
         \mathrm{MPPI}(J(x_0^{(i)},\mathbf{u}))\bigr)$
  \State $\mathcal{D}_\text{batch}\leftarrow\mathcal{D}_\text{batch}\cup\{\tau_i\}$
\EndFor
\State Freeze $\mathcal{D}_\text{batch}$ and train ACT with~\eqref{eq:act_loss}
\State \Return $\theta$
\end{algorithmic}
\end{algorithm}

\begin{figure*}[!t]
  \centering
  \captionsetup[sub]{font=footnotesize, skip=3pt, position=top}
  \begin{subfigure}[t]{0.325\textwidth}
    \centering
    \caption{Panda pick-and-place}
    \vspace{-0.5em}
    \includegraphics[width=\linewidth]{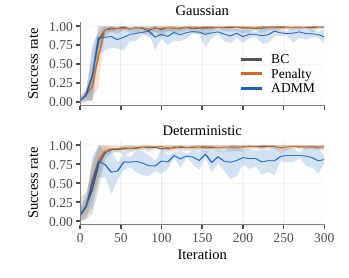}
  \end{subfigure}\hfill
  \begin{subfigure}[t]{0.325\textwidth}
    \centering
    \caption{Panda pick-and-place}
    \vspace{-0.5em}
    \includegraphics[width=\linewidth]{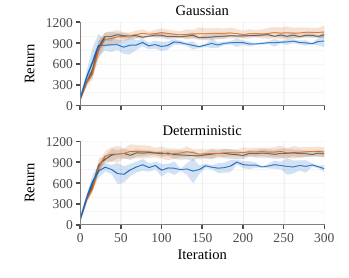}
  \end{subfigure}\hfill
  \begin{subfigure}[t]{0.325\textwidth}
    \centering
    \caption{Go2}
    \vspace{-0.5em}
    \includegraphics[width=\linewidth]{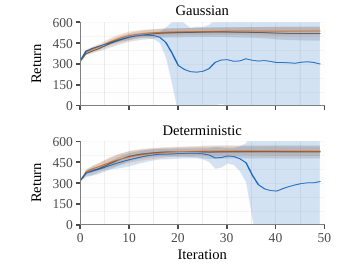}
  \end{subfigure}
  \vspace{-0.5em}
  \caption{The three VAMPS-Iterative coupling strategies. Each task shows a Gaussian policy and a deterministic policy. Shaded bands: 95\,\% CI over 4 seeds.}
  \label{fig:q1_coupling}
  \vspace{-0.8em}
\end{figure*}

\subsubsection{Critic-Guided Policy Learning}\label{sec:iql}
At each iteration, we add the refined MPPI trajectories and the policy rollouts used to warm-start MPPI to a replay buffer $\mathcal{D}$. We use both sources for two reasons. First, the warm-start rollouts provide transitions from the current policy, reducing extrapolation when we evaluate $Q_\phi(x,\pi_\theta(x))$. Second, combining them with the refined MPPI trajectories improves the action coverage used to estimate the advantages.

We train $V_\psi$ and $Q_\phi$ with the IQL losses in~\eqref{eq:iql_v}-\eqref{eq:iql_q}. The buffer $\mathcal{D}$ is used only for these critic updates. AWR~\cite{peng2021advantageweighted} converts the advantage $A_\phi(x,u)=Q_\phi(x,u)-V_\psi(x)$ into the normalized weight
\begin{equation}\label{eq:awr_weight}
  w_\beta(x,u)=
  \frac{\exp\!\left(A_\phi(x,u)/\beta\right)}
       {\E_{(x',u')\sim\mathcal{G}^k}\!\left[\exp\!\left(A_\phi(x',u')/\beta\right)\right]},
\end{equation}
where $\beta>0$ is the temperature. All three coupling strategies use these weights to update the policy on the current MPPI trajectories $\mathcal{G}^k=\{(x_t^{(i)},u_t^{(i),k+1})\}_{i,t}$:
\begin{equation}\label{eq:vamps_policy_update}
\theta^{k+1} = \operatorname*{arg\,min}_\theta
\sum_{(x,u)\in\mathcal{G}^k}
w_\beta(x,u)\,d_\theta(x,u,y).
\end{equation}
BC and Penalty set $y=0$ in this update. ADMM uses its current scaled dual variable $y^k$. After the policy update, ADMM updates this variable at each time step:
\begin{equation}\label{eq:admm_dual}
  y_t^{(i),k+1} \leftarrow y_t^{(i),k}
    + u_t^{(i),k+1}
    - \mu_{\theta^{k+1}}\!\left(x_t^{(i)}\right),
\end{equation}
where $\mu_\theta(x)=\mathbb{E}_{u\sim\pi_\theta(\cdot\mid x)}[u]$ for a Gaussian policy and $\mu_\theta(x)=\pi_\theta(x)$ for a deterministic policy. The dual variable accumulates the planner-policy residual.

\subsection{VAMPS-Batch}\label{sec:vamps_batch}
VAMPS-Iterative uses a one-step proprioceptive policy. Small prediction errors can move this policy away from the states in a fixed training dataset and compound over time~\cite{dagger_2011}. Our frozen-data baseline show this effect in \secref{sec:exp_q7}. VAMPS-Iterative addresses it by repeatedly refining the policy trajectories with MPPI and using them as new supervision. VAMPS-Batch instead uses ACT~\cite{Zhao-RSS-23}, which predicts an action trajectory rather than one action at each step. This reduces the effective decision horizon and motivates training the visuomotor policy once on a fixed dataset.

VAMPS-Batch uses MPPI to autonomously collect this dataset on the real robot. MPPI evaluates candidate rollouts in simulation using task-specific state estimates and executes the refined controls on the robot. For example, it uses the cube position for pick-and-place and the contact height $z$ for erasing. During execution, we record synchronized RGB images, robot observations, task sensors and actions. We freeze the resulting episodes and train ACT offline with~\eqref{eq:act_loss}. At deployment, ACT does not use the task-specific state estimates. Its evaluation is reported in \secref{sec:visuomotor_policy}.

\subsection{Algorithm Summary}\label{sec:algorithm_overview}
Algorithms~\ref{alg:vamps_iterative} and~\ref{alg:vamps_batch} summarize the two VAMPS modes. Both use MPPI to generate supervision from a model and a task cost. VAMPS-Iterative updates the planner, policy and value functions together: the current policy warm-starts MPPI, and each refined trajectory updates the policy before the next outer iteration. VAMPS-Batch instead completes data collection first, freezes the recorded real-robot episodes and then trains ACT offline.

\section{VAMPS-Iterative Experiments \& Results}\label{sec:experiments}

\subsection{Experimental Setup}\label{sec:experimental_setup}

\textbf{VAMPS tasks.}
In Panda pick-and-place from MuJoCo Playground~\cite{zakka2025mujocoplayground}, a Franka Panda grasps a cube from randomized positions and lifts it to a target. We use this task for the coupling comparison, component ablations and standard baselines. Go2 evaluates the coupling on locomotion and real-robot policy transfer.

\textbf{VAMPS-Iterative architecture.}
The policy $\pi_\theta$, critic $(V_\psi,Q_\phi)$ and terminal value $\hat{V}_\omega$ use BroNet~\cite{nauman2024bigger}: a Dense $\to$ LayerNorm $\to$ ReLU stem and residual blocks of width 512. The critic uses two blocks, while $\pi_\theta$ and $\hat{V}_\omega$ use one. $Q_\phi$ is an ensemble of 10 networks~\cite{Chang2026TheSD}, whose mean defines the advantage. The policy output layer uses orthogonal initialization with scale $0.01$. We train all networks with AdamW~\cite{loshchilov2018decoupled}.

\textbf{VAMPS-Iterative training.}
At each outer iteration, the critic and terminal value take 15 gradient steps with batch sizes 1024 and 256, respectively; the terminal value uses $n{=}2$-step targets. We use IQL expectile $\tau{=}0.7$, discount $\gamma{=}0.99$, and target update rate $\tau_\text{net}{=}0.005$. The policy takes 10 epochs on Panda and 15 on Go2, with learning rate $5{\times}10^{-4}$ and AWR temperature $\beta{=}2.0$. MPPI uses horizon $T{=}16$, DIAL-MPC~\cite{Xue2024FullOrderSM} annealing, and replans every two steps. Panda uses $N{=}80$, $K{=}512$, $N_\text{diffuse}{=}3$, and 300 outer iterations. Go2 uses $N{=}50$, $K{=}1024$, $N_\text{diffuse}{=}2$, 50 outer iterations, and domain randomization.

\textbf{VAMPS-Iterative metrics.}
The primary metric is the mean return evaluated on a test set of 50 initial states per seed; shaded bands indicate 95\,\% confidence intervals across seeds.

\textbf{Default configuration.}
Unless stated otherwise, the experiments use VAMPS-Iterative (Algorithm~\ref{alg:vamps_iterative}) with a Gaussian policy.

\subsection{Q1: How does the coupling strategy between the trajectory optimizer and the policy affect performance?}\label{sec:exp_q1}

\textbf{Setup.}
We run the full method with each of the three coupling strategies of \secref{sec:coupling}, and we do so twice: once with a Gaussian policy and once with a deterministic one. This gives six conditions per task, and we run each of them with four seeds. We denote the configurations VAMPS-Iterative (BC), VAMPS-Iterative (Penalty) and VAMPS-Iterative (ADMM). BC uses $\rho=0$, Penalty uses $\rho=0.005$ with the dual variables frozen, and ADMM uses $\rho=0.005$ with the dual variables active.

\textbf{Result.}
On the Panda task, BC and Penalty pass a success rate of $0.90$ within about $30$ iterations and then hold a plateau just under $1.0$, with both the Gaussian and the deterministic policy (\figref{fig:q1_coupling}). ADMM climbs as fast at first but settles lower, at $0.88$ with the Gaussian policy and $0.90$ with the deterministic one. The return shows the same: BC and Penalty plateau between $1010$ and $1060$, with Penalty above BC by about $40$ with both policies. ADMM plateaus between $860$ and $875$. The Go2 task gives the same result. BC and Penalty again behave identically. ADMM is again the worst, and its instability is more visible here. The Go2 curves start from a high return because of the orthogonal action head initialized at a scale of $0.01$: the untrained network outputs a near-zero action, which under position control is the default joint configuration.

\begin{figure}[!tb]
    \centering
    \includegraphics[width=\columnwidth]{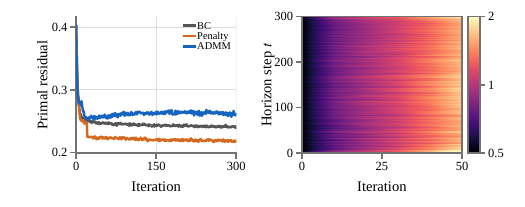}
    \caption{\textbf{Left:}~median primal residual over four seeds: BC already follows the planner to $0.24$ with no consensus enforced. \textbf{Right:}~95th percentile of $|y_{t,j}^{(i)}|$ over states and action coordinates, on one Go2 seed. Colorbar in action units; $1$ is the bound of $\pi_\theta$.}
    \label{fig:q1_analysis}
    \vspace{-1em}
\end{figure}

\textbf{Analysis.}
We attribute the instability of the ADMM variant to the dual variables $\mathbf{y}$. In the policy step~\eqref{eq:vamps_policy_update}, they shift the regression target from $\mathbf{u}$ to $\mathbf{u}+\mathbf{y}$. The right panel of \figref{fig:q1_analysis} shows, at each outer iteration and horizon step $t$, the 95th percentile of $|y_{t,j}^{(i)}|$ over the $N$ initial states and the action coordinates. A cell past $1$ means at least $5\%$ of those entries put $\mathbf{u}+\mathbf{y}$ outside $[-1,1]$, the range of $\pi_\theta$.

With $\mathbf{y}$ set to zero, Penalty at $\rho{=}0.005$ brings no gain over BC, and it brings no loss either. This is consistent with \secref{sec:exp_q8}, where performance degrades as $\rho$ grows. We hypothesize the size of the policy explains this, and the left panel of \figref{fig:q1_analysis} supports it. The coupling term exists to keep the planner inside the set of trajectories that the policy is able to reproduce, which is a real constraint when the policy is represented by a low-capacity neural network, as in GPS~\cite{gpslevine}. Our policy reproduces the MPPI controls with no coupling. The primal residual falls to $0.24$ within $30$ iterations and stays there, and the penalty only brings it to $0.21$. The consensus constraint is therefore almost satisfied. The dual updates move it in the opposite direction, to $0.26$. The displaced target of \figref{fig:q1_analysis} pulls the planner and the policy apart rather than together, which is the behavior the method is meant to prevent. We keep Penalty as the proposed VAMPS-Iterative coupling and use it in the remaining experiments.

\subsection{Q2: Does the learned terminal value function compensate for a shorter horizon?}\label{sec:exp_q5}
\textbf{Setup.}
We evaluate the terminal value $\hat{V}_\omega$ trained at each iteration on the MPPI transitions accumulated in $\mathcal{B}$. It accounts for the part of the task that lies beyond the MPPI horizon, so we vary the horizon as well: $T=16$, $T=8$ and $T=4$, each with and without $\hat{V}_\omega$, for four seeds each.

\begin{figure}[!t]
    \centering
    \includegraphics[width=\columnwidth]{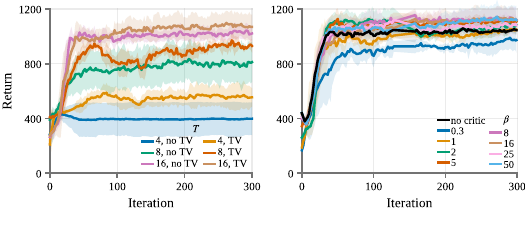}
    \caption{Test return on Panda pick-and-place. \textbf{Left:}~planning horizons $T\in\{16,8,4\}$ with and without the learned terminal value (TV). \textbf{Right:}~VAMPS-Iterative without a critic and with the critic for different AWR temperatures $\beta$; both retain the terminal value. Shaded bands: 95\,\% CI across seeds.}
    \vspace{-0.5em}
    \label{fig:q5_value}\label{fig:q3_critic}
\end{figure}

\textbf{Result.}
As expected, the terminal value improves the final policy return at every horizon, and the gain grows as the horizon shrinks (\figref{fig:q5_value}, left). It is large at $T{=}4$, still clear at $T{=}8$, and small at $T{=}16$. The terminal value is not a replacement for a long horizon, because $T{=}4$ with it stays well below $T{=}8$ without it. The figure therefore shows that a reasonable horizon is important, and that the terminal value lets us reduce the computational budget of MPPI by shortening the rollout.

\subsection{Q3: Does the offline RL critic improve performance?}\label{sec:exp_q6}

\textbf{Setup.}
We compare VAMPS-Iterative with a no-critic ablation, which sets the weights in~\eqref{eq:vamps_policy_update} to one and therefore fits the policy by uniform behavioral cloning on Panda pick-and-place. Both use a Gaussian policy and retain the learned terminal value. We use four seeds per condition. We sweep the AWR temperature $\beta$ over the values in \figref{fig:q3_critic}.

\textbf{Result.}
VAMPS improves slightly over VAMPS-no-critic for the tested temperatures from $\beta{=}2$ to $\beta{=}50$. At $\beta{=}1$ the final return is similar to VAMPS-no-critic, while $\beta{=}0.3$ gives a lower return. The offline RL critic therefore brings a small gain on Panda pick-and-place for a suitable temperature.

\subsection{Q4: How sensitive is performance to the penalty coefficient and batch size?}\label{sec:exp_q8}

\begin{figure}[!t]
    \centering
    \includegraphics[width=\columnwidth]{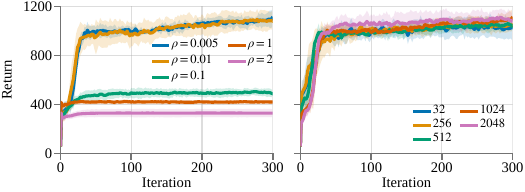}
    \caption{Sensitivity to the hyperparameters. \textbf{Left:}~penalty coefficient $\rho$, with the batch size fixed at $1024$. \textbf{Right:}~batch size, with $\rho$ fixed at $0.005$. Shaded bands: 95\,\% CI over 4 seeds.}
    \vspace{-0.5em}
    \label{fig:q10_hyper}
    \vspace{-1em}
\end{figure}

\textbf{Setup.}
We vary the penalty coefficient $\rho$ and the batch size, one at a time, over the values shown in the legend of \figref{fig:q10_hyper}. All other hyperparameters keep their values from \secref{sec:experimental_setup}. All runs use a Gaussian policy.

\textbf{Result.}
The batch size has no effect, since the five curves overlap over the whole run (\figref{fig:q10_hyper}). The penalty coefficient behaves as a threshold rather than as a gradual trade-off. Values of $\rho=0.005$ and $\rho=0.01$ give the same return, while $\rho\geq 0.1$ drops to a much lower level within the first iterations and stays there. The policy is untrained at the beginning, so a large $\rho$ makes the penalty dominate the MPPI cost.

\subsection{Q5: How do standard baselines perform in this regime?}\label{sec:exp_q7}

\textbf{Setup.}
We compare VAMPS-Iterative against PPO~\cite{schulman2017proximalpolicyoptimizationalgorithms}, frozen BC and frozen IQL on the Panda pick-and-place task, with four seeds and $50$ test initial states. PPO learns from the same reward. We train it at three episode lengths: $80$, which matches VAMPS-Iterative, and $500$ and $1000$. Evaluation is always $80$ steps. BC and IQL learn offline from one pass of MPPI transitions. We give PPO the same simulator budget as VAMPS-Iterative, $1.6\!\times\!10^{10}$ steps. Of that budget VAMPS-Iterative uses only $1.3\!\times\!10^{6}$ policy--environment interaction steps, in episodes of length $80$, the rest are planner rollouts, and those policy steps go into the critic's replay buffer. Every PPO step is a policy-environment interaction step.

\begin{table}[t]
    \centering
    \caption{Baselines on Panda: mean $\pm$ 95\,\% CI over four seeds at each method's best checkpoint. PPO is trained at the listed episode length and evaluated over $80$ steps, as VAMPS-Iterative.}
    \label{tab:baselines}
    \small
    \setlength{\tabcolsep}{5pt}
    \begin{tabular}{llrr}
    \toprule
    Method & Episode length & Return  & Success \\
    \midrule
    PPO             & $80$   & $279 \pm 84$  & $0.00$ \\
    PPO             & $500$  & $975 \pm 10$  & $\textbf{0.98} \pm \textbf{0.03}$ \\
    PPO             & $1000$ & $894 \pm 163$ & $0.74 \pm 0.49$ \\
    BC (frozen)     & --    & $290 \pm 85$  & $0.02 \pm 0.03$ \\
    IQL (frozen)    & --    & $240 \pm 40$  & $0.01 \pm 0.01$ \\
    VAMPS-Iterative & $80$   & $\textbf{1109} \pm \textbf{76}$ & $\textbf{0.98} \pm \textbf{0.02}$ \\
    \bottomrule
    \end{tabular}
\end{table}

\textbf{Result.}
Table~\ref{tab:baselines} shows that PPO trained at episode length $80$ stays at $279$ return and $0.00$ success where the gripper moves toward the cube and never picks it up. Increasing the training episode to $500$ raises the mean to $975$ / $0.98$; at $1000$ the mean is $894$ / $0.74$, pulled down by one seed that also never grasps, and stays below VAMPS-Iterative. Frozen BC and IQL are trained on one pass of successful MPPI trajectories, yet they stay with PPO at episode length $80$ ($290$ / $0.02$ and $240$ / $0.01$). VAMPS-Iterative, at episode length $80$, reaches $1109$ / $0.98$. It passes a return of $1000$ after $3.6\!\times\!10^{9}$ simulator steps, of which only $2.9\!\times\!10^{5}$ are policy-environment interaction steps used to warm-start the planner.

The key difference is exploration. At $50$\,Hz, VAMPS-Iterative uses $80$-step ($1.6$\,s) episodes, while PPO needs $500$-step ($10$\,s) episodes of an untrained policy to solve the task, which is unsafe on hardware. VAMPS-Iterative instead explores with MPPI on a model. At every iteration, the current policy warm-starts MPPI, which refines its actions and returns new trajectories to update the policy. This feedback loop focuses planning around the current policy and provides new supervision as the policy changes. Frozen BC compounds errors outside the MPPI states~\cite{dagger_2011}, while frozen IQL has no coverage beyond its single pass of $6{,}320$ transitions~\cite{kostrikov2022offline,Levine2020OfflineRL}.

\subsection{Q6: Does the policy transfer to real hardware?}\label{sec:exp_go2_deploy}
We deploy the proprioceptive policy trained in simulation on a physical Unitree Go2 robot. \figref{fig:go2_deploy} shows four frames of one bounding cycle recorded during such a run, where the robot moves under the commanded velocity from the joystick. We used domain randomization~\cite{Tobin2017DomainRF}. Each of the $N$ initial states carries its own randomized model, drawn once before training and then held fixed for the whole run. We randomize the friction of the floor, the mass of every link, a payload of up to one kilogram on the torso, the position of the torso center of mass, the joint friction and armature, and the PD gains of the position actuators. MPPI plans in the perturbed model of its own scenario, so the policy is fitted on solutions that already account for the model mismatch.

\section{VAMPS-Batch Experiments \& Results}\label{sec:visuomotor_policy}
Both VAMPS-Batch experiments run on a 7-DOF Flexiv Rizon 10S with a gripper and a side-mounted Orbbec Femto Bolt camera. MPPI evaluates $2048$ sampled rollouts in simulation over a horizon $T{=}20$. It replans in receding horizon and executes the refined controls on the robot at $25$\,Hz through an impedance controller. We record RGB images, sensor observations and actions, and train ACT on these real-robot episodes using the HuggingFace LeRobot implementation~\cite{cadene2024lerobot,Zhao-RSS-23}. At inference, the full action trajectory is executed open-loop before ACT is queried again.
\subsection{Experiment 1: Visuomotor pick-and-place}

\textbf{Setup.}
The robot grasps a cube from a randomized position on the table and drops it in a fixed bowl. We localize the cube with HSV color filtering and a calibrated homography. This estimate is used only by MPPI during data collection. We autonomously collect 100 MPPI episodes. The policy predicts a chunk of $T_u{=}100$ joint-position targets.

\textbf{Result.}
The VAMPS-Batch policy succeeds in 8 of 10 trials from randomized cube positions. The middle row of \figref{fig:go2_deploy} shows four frames of one successful execution.

\begin{figure}[!t]
    \centering
    \includegraphics[width=\columnwidth]{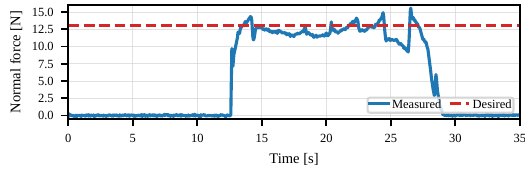}
    \caption{Desired and measured normal force during one successful VAMPS-Batch erasing trial shown in \figref{fig:go2_deploy}.}
    \vspace{-1em}
    \label{fig:act_erasing_force}
    \vspace{-1em}
\end{figure}

\subsection{Experiment 2: Visuomotor and force-aware erasing policy}

\textbf{Setup.}
The robot erases a whiteboard placed on the table. The eraser has to stay in contact with the board while the arm moves over it, so MPPI needs the height at which to hold the eraser: too high and it loses contact and does not erase, too low and it presses through the board. During data collection, we obtain this height with the task-space impedance-admittance nonlinear MPC of~\cite{nguyen2026taskspace}. It regulates the contact and gives the $z$ position MPPI plans around. This contact-height estimate is used only during data collection. We collect demonstrations at discrete desired normal-force setpoints from $5$ to $14$\,N to cover different contact conditions. The resulting dataset contains 84 episodes ($1.59$ hours on the robot). For this task, ACT also takes a second camera mounted on the wrist. Its observation at time $t$ contains the two synchronized images and a $14$-D vector with the seven joint positions, the measured $6$-D wrench and the desired normal force. ACT reads the wrench at $t$ without averaging a force window. The policy predicts a chunk of $T_u{=}50$ joint-position targets ($2$\,s at $25$\,Hz), executed open-loop as in pick-and-place.

\textbf{Result.}
The same VAMPS-Batch checkpoint succeeds in four consecutive trials and erases the writing in all of them. \figref{fig:act_erasing_force} shows the desired and measured normal force during one successful trial. The bottom row of \figref{fig:go2_deploy} shows four frames of the same execution. These preliminary results show that VAMPS-Batch can train visuomotor policies from autonomously collected real-robot data without human demonstrations. We plan to extend this mode to more sensor-based tasks in future work.

\section{Conclusions}\label{sec:conclusion}

We presented VAMPS, a framework that uses a sampling-based planner to train reusable robot policies without human demonstrations or a differentiable simulator. We studied two modes of this framework. VAMPS-Iterative alternates MPPI trajectory refinement and policy learning for one-step proprioceptive policies. Our experiments show that iterative refinement outperforms training once on frozen MPPI data, simple planner-policy coupling is sufficient, and a learned terminal value improves short-horizon planning. We transfer the resulting locomotion policy to a Unitree Go2. VAMPS-Batch uses MPPI to autonomously collect a fixed real-robot dataset and trains chunked visuomotor policies offline with ACT. The dataset contains RGB images, robot observations, task sensors and actions. We demonstrate this mode on pick-and-place and force-aware whiteboard erasing with a Flexiv Rizon 10S.

The VAMPS-Batch hardware results are preliminary, and we plan to extend it to more sensor-based tasks. For long-horizon tasks, TAMP~\cite{garrett2020integratedtaskmotionplanning} could provide a high-level planner. Scene generators such as SimFoundry~\cite{ranawaka2026simfoundrymodularautomatedscene} could build simulations from real footage, allowing VAMPS to collect data and learn visuomotor policies in a real-to-sim-to-real loop. Future work will also study tighter planner--policy feedback in architectures such as TD-MPC2~\cite{hansen2024tdmpc2scalablerobustworld} and CAST~\cite{crestaz_CAST}.

\bibliographystyle{ieeetr-mod}

\bibliography{references.bib}

\end{document}